\documentclass{article}

\usepackage{arxiv}

\usepackage{subfig}
\usepackage{mlmath}
\usepackage{amsthm}

\usepackage[utf8]{inputenc} 
\usepackage[T1]{fontenc}    
\usepackage{hyperref}       
\usepackage{url}            
\usepackage{booktabs}       
\usepackage{amsfonts}       
\usepackage{nicefrac}       
\usepackage{microtype}      
\usepackage{lipsum}
\usepackage{graphicx}
\graphicspath{ {./images/} }

\newtheorem{theorem}{Theorem}
\newtheorem{lemma}{Lemma}

\title{Loss Jump During Loss Switch in Solving PDEs with Neural Networks}

\author{
 Zhiwei Wang \\
  Institute of Natural Sciences, School of Mathematical Sciences\\
  Shanghai Jiao Tong University\\
  Shanghai 200240, P.R. China \\
  \texttt{victorywzw@sjtu.edu.cn} \\
   \And
 Lulu Zhang \\
  Institute of Natural Sciences, School of Mathematical Sciences\\
  Shanghai Jiao Tong University\\
  Shanghai 200240, P.R. China \\
  \texttt{zhangl9661@sjtu.edu.cn} \\
  \And
 Zhongwang Zhang \\
  Institute of Natural Sciences, School of Mathematical Sciences\\
  Shanghai Jiao Tong University\\
  Shanghai 200240, P.R. China \\
  \texttt{0123zzw666@sjtu.edu.cn} \\
  \And
 Zhi-Qin John Xu \\
  Institute of Natural Sciences, School of Mathematical Sciences\\
  Shanghai Jiao Tong University\\
  Shanghai 200240, P.R. China \\
  \texttt{xuzhiqin@sjtu.edu.cn} \\
}

\begin{document}
\maketitle
\begin{abstract}
Using neural networks to solve partial differential equations (PDEs) is gaining popularity as an alternative approach in the scientific computing community. Neural networks can integrate different types of information into the loss function. These include observation data, governing equations, and variational forms, etc. These loss functions can be broadly categorized into two types: observation data loss directly constrains and measures the model output, while other loss functions indirectly model the performance of the network, which can be classified as model loss. However, this alternative approach lacks a thorough understanding of its underlying mechanisms, including theoretical foundations and rigorous characterization of various phenomena. This work focuses on investigating how different loss functions impact the training of neural networks for solving PDEs. We discover a stable loss-jump phenomenon: when switching the loss function from the data loss to the model loss, which includes different orders of derivative information, the neural network solution significantly deviates from the exact solution immediately. Further experiments reveal that this phenomenon arises from the different frequency preferences of neural networks under different loss functions. We theoretically analyze the frequency preference of neural networks under model loss. This loss-jump phenomenon provides a valuable perspective for examining the underlying mechanisms of neural networks in solving PDEs.
\end{abstract}


\section{Introduction}\label{sec:Introduction}

The use of neural networks for solving partial differential equations (PDEs) has emerged as a promising alternative to traditional numerical methods in the scientific computing community. By incorporating various types of information into the loss function, such as observation data, governing equations, and variational forms, neural networks offer a flexible and powerful framework for approximating the solution of PDEs. These loss functions can be broadly classified into two categories: data loss, which directly constrains and measures the model output using observation data, and model loss, which indirectly models the performance of the network using equations and variational forms.

Despite the growing interest in this approach, a comprehensive understanding of the underlying mechanisms governing the behavior of neural networks in solving PDEs is still lacking. While several works have explored the capabilities and limitations of physics-informed learning \cite{xu2019frequency,karniadakis2021physics,mishra2022estimates,duan2022convergence,jiao2022rate} and the challenges in training physics-informed neural networks (PINNs) \cite{xu2019frequency,wang2022and}, the impact of different loss functions on the training dynamics and convergence properties of neural networks remains an open question.

Recent studies have shown that the derivatives of the target functions in the loss function play a crucial role in the convergence of frequencies \cite{xu2019frequency,lu2021deepxde,xu2022overview}. A key observation is that neural networks often exhibit a frequency principle, learning from low to high frequencies \cite{xu2019frequency,xu2019training,rahaman2019spectral,xu2022overview}. This phenomenon has inspired a series of theoretical works aimed at understanding the convergence properties of neural networks \cite{luo2019theory,luo2022exact,basri2019convergence,cao2019towards,bordelon2020spectrum}.

Moreover, the development of deep learning theory and algorithms has greatly benefited from the accurate description of stable phenomena. For instance, it has been observed that heavily over-parameterized neural networks usually do not overfit \cite{zhang2016understanding,breiman1995reflections}, neurons in the same layer tend to condense in the same direction \cite{luo2021phase,zhou2022towards,zhang2024implicit}, and stochastic gradient descent or dropout tends to find flat minima \cite{keskar2016large,wu2018sgd,zhu2019anisotropic,smith2020origin,feng2021inverse,zhang2024implicit}. Additionally, a series of multiscale neural networks have been developed for solving differential equations \cite{liu2020multi,li2023subspace,jagtap2020adaptive,wang2021eigenvector,zhang2022multi,wang2024deep} and fitting functions \cite{tancik2020fourier,mildenhall2020nerf}.


Motivated by these findings, we aim to investigate the impact of different loss functions on the training dynamics and convergence properties of neural networks for solving PDEs. We focus on the interplay between data loss and model loss, which incorporate different orders of derivative information. We discover a stable loss-jump phenomenon: when switching the loss function from the data loss to the model loss, which includes different orders of derivative information, the neural network solution significantly deviates from the exact solution immediately. 

In this work, we analyze the training process and the dynamics induced by different loss functions from a frequency-space perspective. We identify a multi-stage descent phenomenon, where the neural network's ability to constrain low-frequency information is weak when using model loss functions. Furthermore, we model the training process of two-layer neural networks with two-order derivative loss and quantitatively prove that within a certain frequency range, neural networks with high-order derivative loss functions are more inclined to fit high-frequency information in the target function.

The insights gained from this study shed light on the complex interplay between loss functions, frequency preference, and convergence properties of neural networks in solving PDEs. By understanding these mechanisms, we aim to contribute to the development of more robust and efficient neural network-based PDE solvers and to advance the theoretical foundations of this rapidly evolving field.

\section{Fitting Target Function with Different Loss Functions}

    Given an objective function $u(\vx, t)$, an appropriate loss function can be selected based on the available data to train a neural network $u_{\vtheta}(\vx, t)$ that approximates the target function.

    \subsection{Data loss}
    
    Assuming a set of $N$ observation points and corresponding values ${((\vx_f^i, t_f^i), u(\vx_f^i, t_f^i))}_{i=1}^N$, the mean squared error between the target function and the neural network output can be used as a loss function to train the network. This data loss is represented as

    \begin{equation}
        L_{\mathrm{data}}(\vtheta ) = \frac{1}{N}\sum_{i=1}^{N}\norm{u_{\vtheta}(\vx_f^i, t_f^i)-u(\vx_f^i, t_f^i)}^2. \label{eq: data loss}  
    \end{equation}

    The data loss serves not only as a loss function for supervised learning but also as a direct measure of the distance between the model's output and the true target function. Consequently, it is the most important metric for evaluating the performance of the algorithm. By minimizing the data loss, the optimal set of parameters $\vtheta$ is sought, enabling the neural network $u_{\vtheta}$ to closely approximate the target function $u$.

    \subsection{Model loss}

    Model loss refers to the use of the governing equations of the target function as a supervised learning metric. By calculating the mean squared loss between the model's predictions and the target function's governing equations, the model loss function is obtained. The governing equations can take various forms, such as combinations of multi-order derivatives, PDE control equations, or variational formulations. Here are a few examples:

    If the first-order derivatives of the function are available at several observation points, they can be used in conjunction with boundary and initial conditions to construct a loss function for training the neural network. Similarly, higher-order derivative information can also be incorporated into the loss function. Such a model loss can be represented as

    \begin{equation}
        L_{\mathrm{model}}(\vtheta) = \sum_{k=0}^n\frac{\lambda_k}{N_k}\sum_{i=1}^{N_k}\norm{u^{(k)}_{\vtheta}(\vx_f^i, t_f^i)-u^{(k)}(\vx_f^i, t_f^i)}^2 \label{eq:pde model loss} 
    \end{equation} 

    where $N_k$ is the number of sampled $k$-order derivative values of the objective function, and $\lambda_k$ is the weight assigned to the corresponding loss term.
    
    If the governing equation of the objective function is known, for example, if the function satisfies a PDE of the form
    
    \begin{align}
        \fL u(\vx, t) &= f(\vx, t), \quad\vx\in\Omega, \ t\in [0, T],\label{eq:pde problem}\\ 
        u(\vx, 0) &= h(\vx),\nonumber \\
        u(\vx, t) &= g(\vx, t), \quad\vx \in \partial\Omega,\nonumber 
    \end{align}
    where $\Omega\subset \sR^d$ denotes the spatial domain, $[0, T]$ is the time interval, $u: \bar{\Omega}\times [0, T]\rightarrow \sR^n$ is the exact solution, $f:\Omega\times [0, T]\rightarrow \sR^n$ is the source term, and $h:\Omega\rightarrow \sR^n$ and $g:\Omega\times [0, T]\rightarrow \sR^n$ are the initial and boundary conditions, respectively, the governing equation can be directly incorporated into the loss function. Specifically, the model loss can include the residual of the governing equation, initial value loss, boundary value loss, and optional supervised learning point loss:
    
    \begin{align}
        L_{model}(\vtheta) &= \frac{\lambda_f}{N_f}\sum_{i=1}^{N_f}\norm{\fL u_{\vtheta} (\vx_f^i, t_f^i)-f(\vx_f^i, t_f^i)}^2 \nonumber \\
                          &+ \frac{\lambda_h}{N_h}\sum_{i=1}^{N_h}\norm{u_{\vtheta} (\vx_h^i, 0) - h(\vx_h^i)}^2 \nonumber \\
                          &+ \frac{\lambda_g}{N_g}\sum_{i=1}^{N_g}\norm{u_{\vtheta} (\vx_g^i, t_g^i)-g(\vx_g^i, t_g^i)}^2 \nonumber \\
                          &+ \frac{\lambda_s}{N_s}\sum_{i=1}^{N_s}\norm{u_{\vtheta} (\vx_s^i, t_s^i)-u(\vx_s^i, t_s^i)}^2,\label{eq:pde model loss2} 
    \end{align}  
    where $N_f$, $N_h$, $N_g$, and $N_s$ are the numbers of collocation points, training points sampled from the initial condition, training points on the boundary, and supervised learning points, respectively. $\lambda_f$, $\lambda_h$, $\lambda_g$, and $\lambda_s$ are the corresponding weight hyperparameters used to balance the contributions of each loss term. It should be noted that the supervised learning point loss is not always necessary, as in most cases, the initial and boundary conditions along with the PDE form are sufficient to determine the solution.

    Another approach is to use the variational form of the governing function as the loss function, which is known as the deep Ritz method. Suppose the variational form satisfied by $u$ is

    \begin{equation}
        I(u) = \int_\Omega (\frac{1}{2}\abs{\nabla u(\vx)}^2-f(\vx)u(\vx))\diff \vx
    \end{equation}

    Then the loss function can be designed as

    \begin{equation}
        L(\vtheta) = \frac{1}{N}\sum_{k=1}^N \frac{1}{2}\abs{\nabla u_\theta(\vx)} - f(\vx)u_\theta(\vx)
    \end{equation}

    Model loss provides a variety of choices for training neural networks and, when used appropriately, can accelerate convergence. However, it cannot directly measure the gap between the model's output and the target function. In some cases, even when the model loss is small, the model's error can be extremely large. For example, in the interval $[-1/\epsilon, 1/\epsilon]$, if $\sin(\epsilon x)$ is trained using the first-order derivative, where $\epsilon$ is sufficiently small, the model loss is $O(\epsilon)$ when the model output is 0, but the data loss is $O(1)$. Therefore, model loss serves as an indirect evaluation metric rather than a direct one.

\section{Rapid Increase in Error When Switching Loss}\label{sec:rapid increase}


    In this work, we discover a loss-jump phenomenon: when switching the loss function from low-order to high-order derivatives, such as switching from data loss to model loss, the neural network solution significantly deviates from the exact solution immediately.

    To illustrate this phenomenon, we begin with a simple Poisson problem:
    
    \begin{align}
        \Delta u(x) &= -\sin(x)-100\sin(10x), \quad x\in[0, 2\pi],\label{eq:toy model}\\ 
        u(0) &= u(2\pi) = 0.\nonumber 
    \end{align}

\begin{figure}[htbp]
        \centering
        \includegraphics[width=\linewidth]{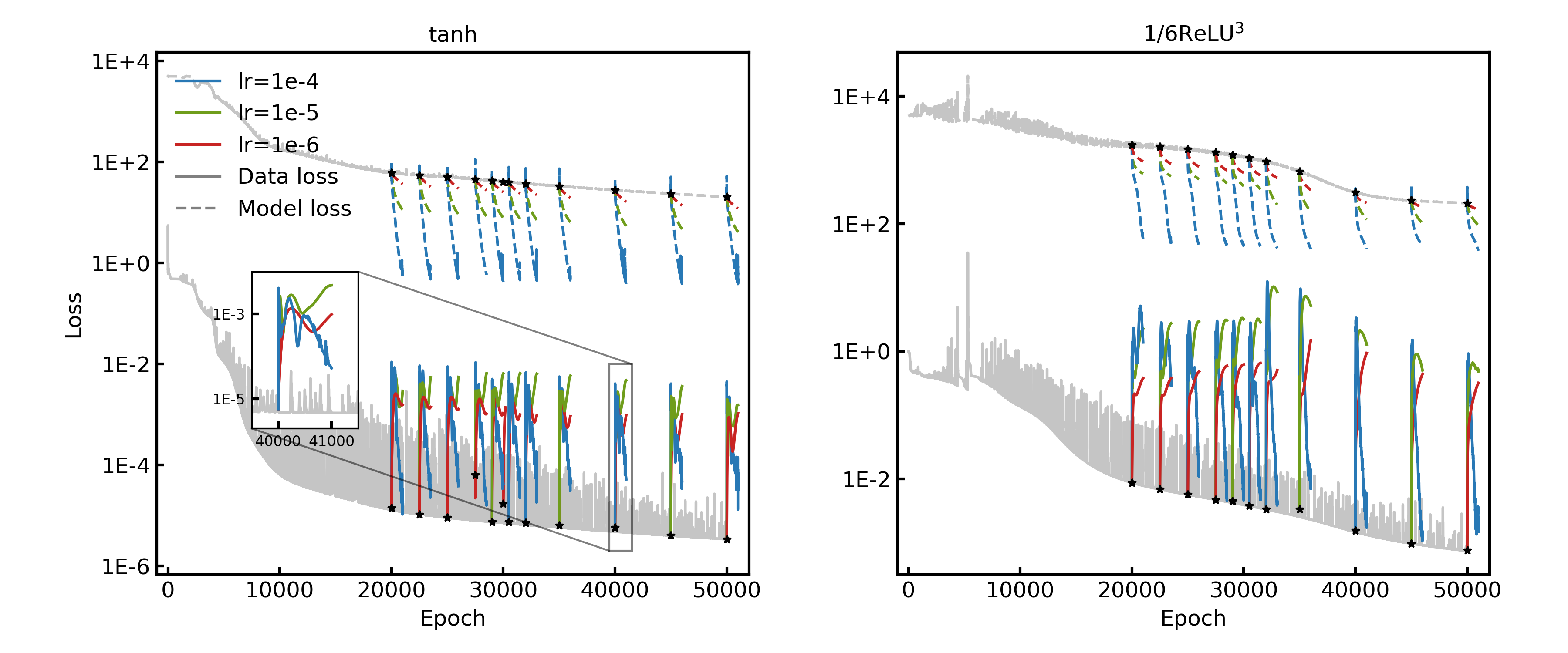}
        \caption{Training process under different learning rate with tanh (left) and ReLU (right) activation function. The gray line indicates pre-training using the data loss function. The asterisk points the error when switching loss. The colored lines are different learning rates used.}
        \label{fig:loss_switch}
    \end{figure}

    The exact solution to this problem is $u(x)=\sin(x)+\sin(10x)$. We selected 5120 equidistant points as training data and employed a neural network with 3 hidden layers, each containing 320 neurons. The commonly used $\tanh$ and cubic ($\frac{1}{6}\ReLU^3$) activation functions were utilized, along with the Adam optimizer \cite{kingma2014adam}. The data loss was trained for 100,000 epochs, and the model loss was introduced at different epochs to observe changes in both data and model losses. The model loss weights $(\lambda_f, \lambda_h, \lambda_g, \lambda_s)$ were set to $(1, 10, 10, 0)$, indicating that no additional supervised data points were used for model loss training. During pre-training, the learning rate was set to 1e-3, decaying to 92\% of its original value every 1000 epochs. Different learning rates were employed after switching the loss function. As shown in Fig. \ref{fig:loss_switch}, the results demonstrate that the increase in data loss after switching the loss function is consistent across different learning rates, indicating that this phenomenon is not caused by an excessively high learning rate.

    The same experimental phenomenon can also be observed in other PDE equations. We examine several types of equations, including the Burgers equation, heat equation, diffusion equation, and wave equation.
    
    The results are listed below. The neural network structure used is a fully connected network with 5 hidden layers, each containing 40 neurons. The $\tanh$ activation function is employed, and the network is initialized using Glorot normal initialization \cite{glorot2010understanding}. The Adam optimizer is used for training. The first 50,000 epochs utilize model loss, with a learning rate of 1e-3 that decays to 92\% of its original value every 1000 epochs. Both the training and test sets are Cartesian products of 500 equidistant space grid points and 11 equidistant time grid points. After switching to the model loss function, training continues for an additional 50,000 epochs. At each epoch, Monte Carlo sampling is used to select 8192 points within the region as the training dataset. The learning rate after switching the loss function is 1e-5 for the Burgers problem and 1e-4 for the others, decaying to 95\% of its original value every 1000 epochs. Additionally, 100 points are selected at both the boundary and initial area for supervised learning. The test set remains the previously defined 5500 equidistant grid points. The weights for each component of the model loss are set to $(\lambda_f, \lambda_h, \lambda_g)=(1,1,1)$.

    \subsection{Burgers equation}

        The Burgers equation, proposed by Dutch mathematician Johannes Martinus Burgers in 1948, is often used to study fluid mechanics, turbulence, and shock waves. The equation captures the key physical processes of fluid motion, including nonlinear convective effects and viscous dissipation. It serves as a simple yet important model for understanding and studying complex fluid problems. The Burgers equation has the following form:
        
        \begin{align}
            u_t + uu_x - \frac{0.01}{\pi}u_{xx} &= 0, \quad x\in[-1, 1], \ t\in [0, 1],\label{eq:Burgers}\\ 
            u(x, 0) &= -\sin(x),\nonumber \\
            u(x, t) &= 0, \quad x=-1, 1.\nonumber 
        \end{align}
    
        \begin{figure}[htbp]
    	\centering
    	\includegraphics[width=\linewidth]{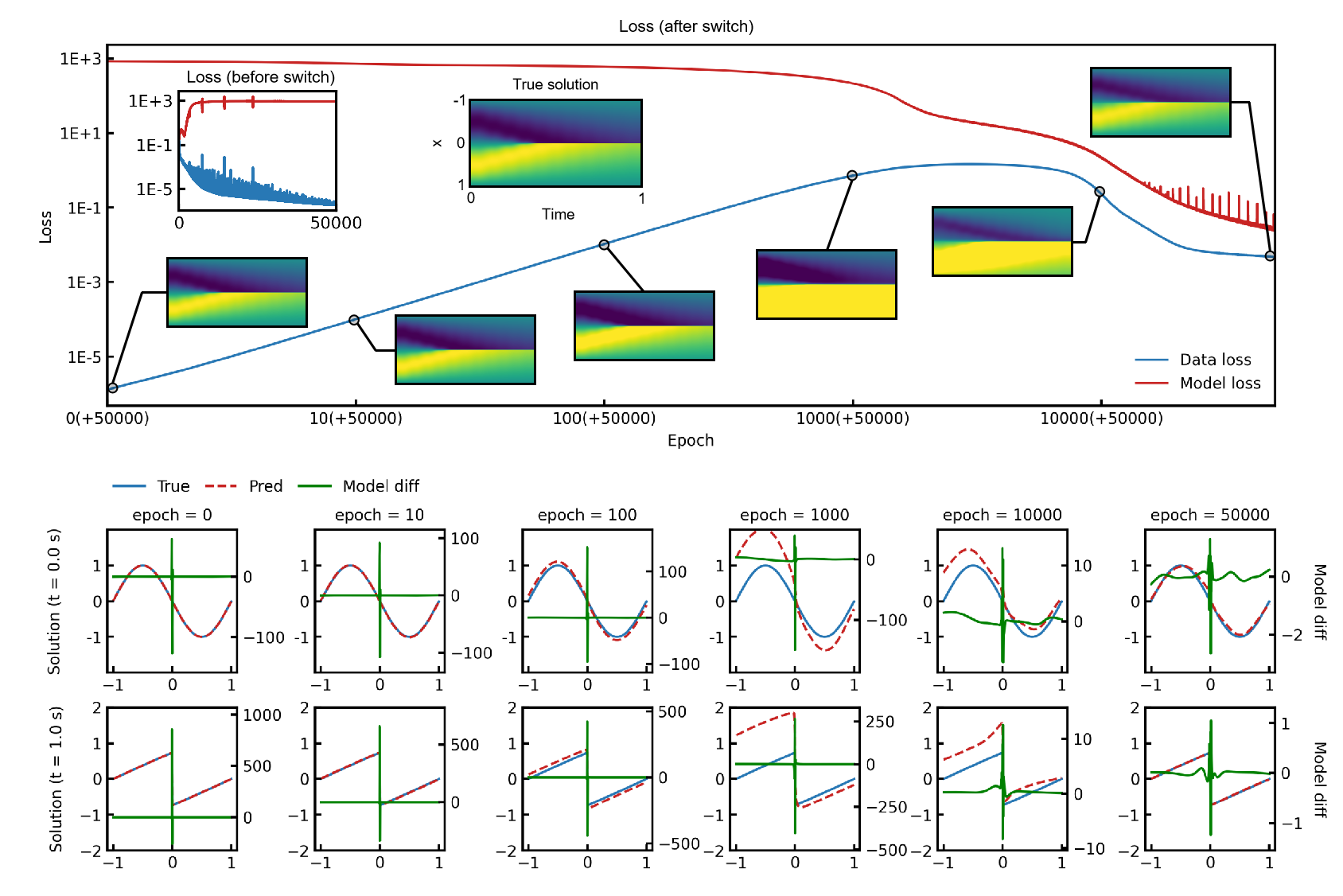}
    	\caption{Burgers equation training process. The second and third rows are the changes of the network prediction value with the training process at time $t=0$ and time $t=1$ respectively after switching the loss.}
    	\label{fig:Burgers_training_process}
        \end{figure}

         Fig. \ref{fig:Burgers_training_process} shows the results after switching to model loss following 50,000 epochs of data loss training. It can be observed that the predictions initially deviate as a whole and then converge to another minimum point.
         
    \subsection{Heat equation}

        The heat equation is a differential equation that describes heat conduction and diffusion. Here, we consider a classic example of the equation, which describes a purely conductive process without a heat source, following the assumption of local thermodynamic equilibrium:

        \begin{align}
            u_t - u_{xx} &= 0, \quad x\in[0, 1], \ t\in [0, 1],\label{eq:heat}\\ 
            u(x, 0) &= \sin(\pi x),\nonumber \\
            u(x, t) &= 0, \quad x=0, 1.\nonumber 
        \end{align}

        And the analytical solution is $u(x, t)=e^{-\pi^2t}\sin(\pi x)$.

        \begin{figure}[htbp]
    	\centering
    	\includegraphics[width=\linewidth]{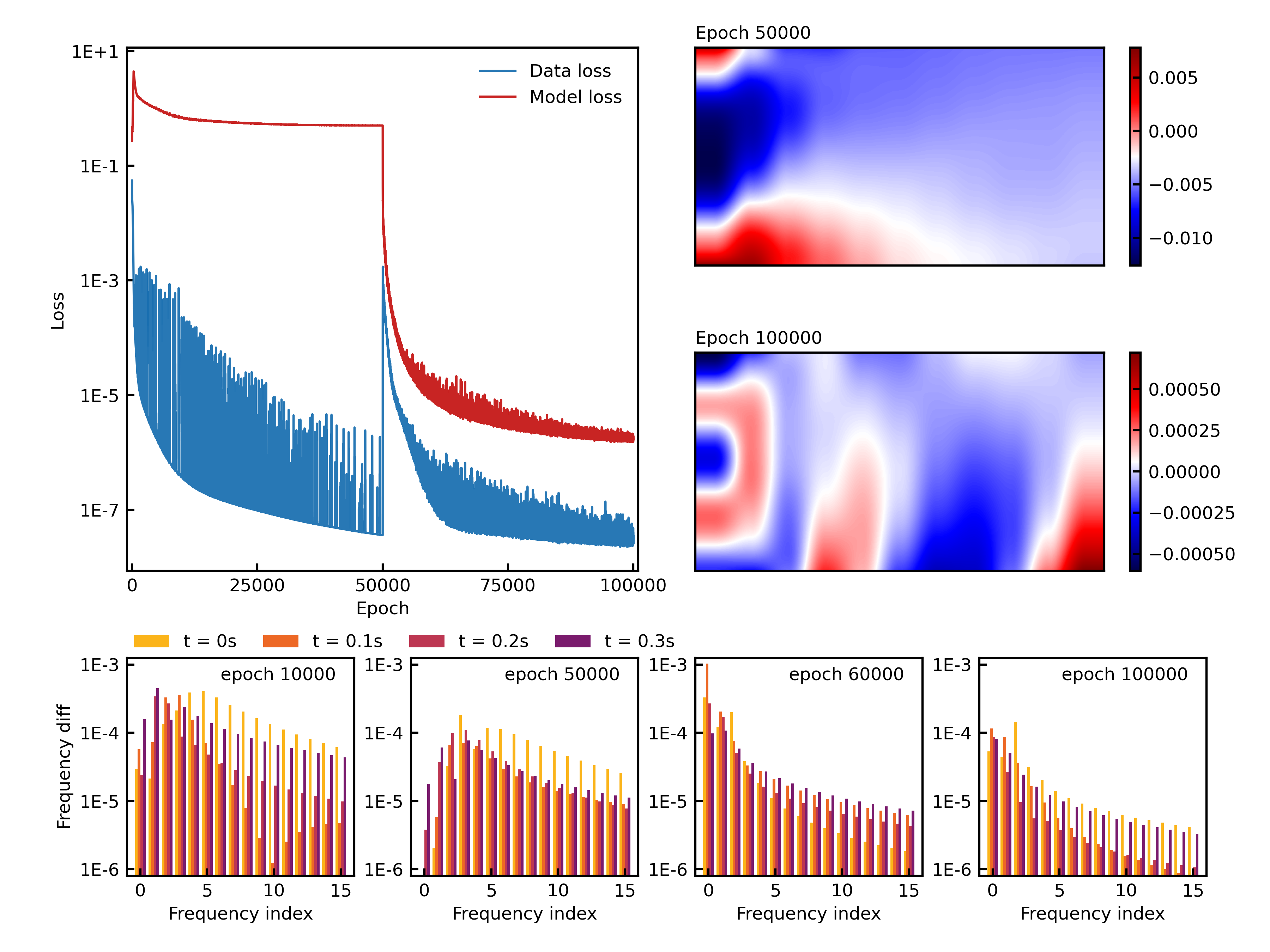}
    	\caption{Heat equation training process. Upper left: data loss and model loss as the training progresses. Upper right: error heatmaps at 50,000 and 100,000 epochs. Bottom: Frequency error at different training epochs. The 4 sub-figures are the results from different training stage.}
    	\label{fig:heat_training_process}
        \end{figure}

        From the loss plot in the upper left corner of Fig.\ref{fig:heat_training_process}, we can see that when switching to model loss after 50,000 epochs, the data loss rises sharply, indicating that the solution has jumped out of the local optimum. It is important to note that the model loss drops after switching, suggesting that this phenomenon is not caused by the shock of an excessively large learning rate. In fact, the learning rate of 1e-4 used in this experiment is much smaller than the learning rates typically used in physics-informed neural networks (PINNs). The two error heatmaps on the right illustrate that the minimum points obtained after switching the loss function differ from the original ones. We believe this may be due to the fact that the dynamical behaviors induced by the two loss functions exhibit different frequency preferences. The bottom row of Fig.\ref{fig:heat_training_process} shows the variation of frequency error at different training stages and time slices. In the first 50,000 epochs of training, it can be clearly seen that low-frequency information is fitted first. However, when the loss function is switched, the error shows a decreasing trend with frequency. This may be the main reason for the sudden increase in errors.

        \begin{figure}[htbp]
    	\centering
    	\includegraphics[width=\linewidth]{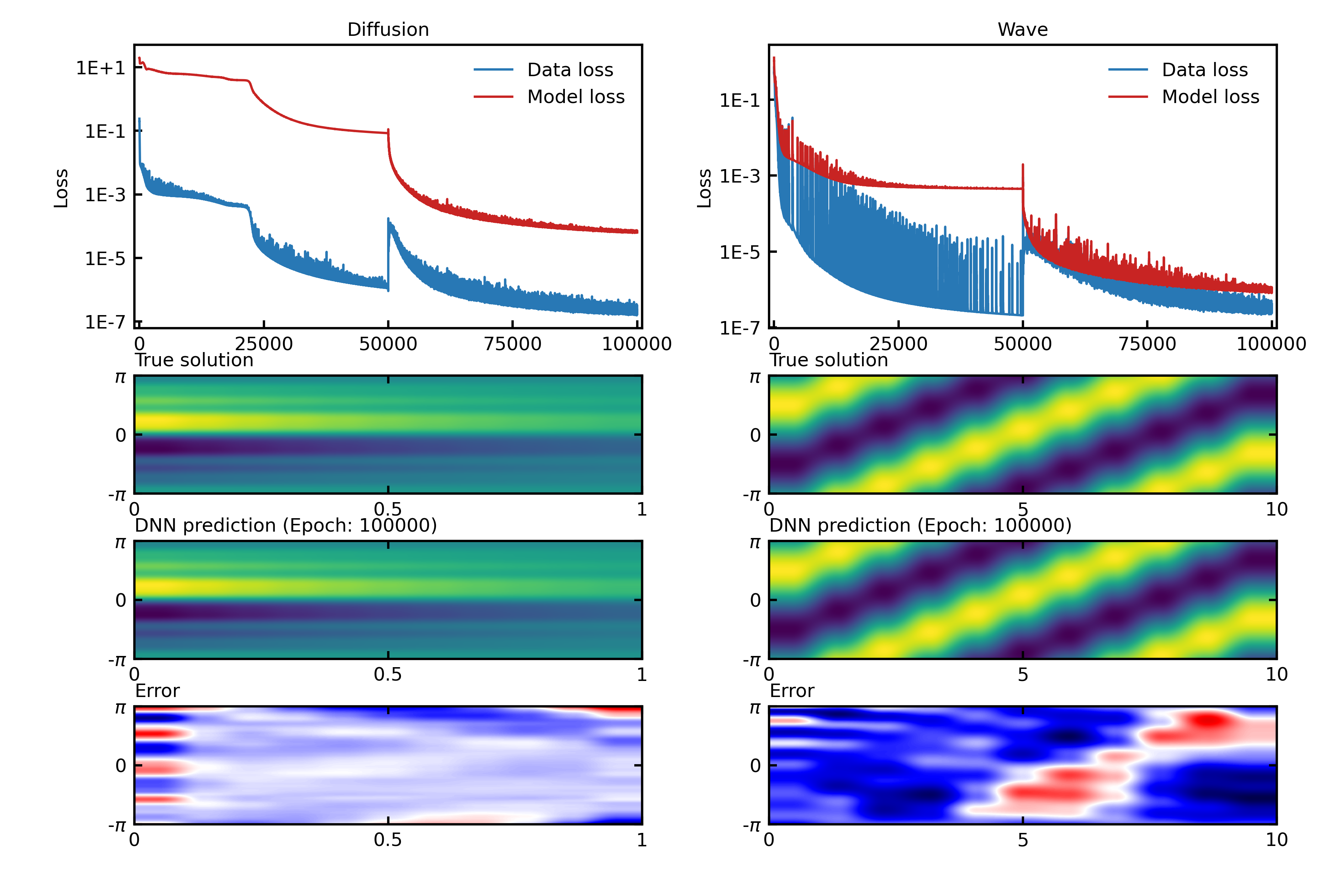}
    	\caption{Diffusion equation (left) and wave equation (right) training process. Top: data loss and model loss as the training progresses. Middle: Heatmap of the analytical solution and the DNN-predicted solution. Bottom: Absolute error between analytical solution and DNN prediction.}
    	\label{fig:diffusion_wave_training_process}
        \end{figure}

    \subsection{Diffusion equation and wave equation}
        We also tested the diffusion equation and the wave equation and obtained similar results.

        For the diffusion equation, we fabricated an analytical solution with decreasing frequency magnitudes. The PDE formulation of the diffusion equation is defined as:
        
        \begin{align}
            u_t - u_{xx} &= e^{-t}(\frac{3}{2}\sin(2x)+\frac{8}{3}\sin(3x)+\frac{15}{4}\sin(4x)+\frac{63}{8}\sin(8x)), \quad x\in[-\pi, \pi], \ t\in [0, 1],\label{eq:diffusion}\\ 
            u(x, 0) &= \sin(x)+\frac{1}{2}\sin(2x)+\frac{1}{3}\sin(3x)+\frac{1}{4}\sin(4x)+\frac{1}{8}\sin(8x),\nonumber \\
            u(x, t) &= 0, \quad x=-\pi, \pi.\nonumber 
        \end{align}

        The exact solution is $u(x, t)=e^{-t}(\sin(x)+\frac{1}{2}\sin(2x)+\frac{1}{3}\sin(3x)+\frac{1}{4}\sin(4x)+\frac{1}{8}\sin(8x))$.
        
        For the wave equation, we used an analytical solution with only a single frequency. The wave equation is defined as:

        \begin{align}
            u_{tt} - u_{xx} &= 0, \quad x\in[-\pi, \pi], \ t\in [0, 10],\label{eq:wave}\\ 
            u(x, 0) &= \sin(x),\nonumber \\
            u(x, t) &= \sin(t), \quad x=-\pi, \pi.\nonumber 
        \end{align}

        The exact solution is $u(x, t)=\sin(x-t)$.

        \begin{figure}[htbp]
    	\centering
    	\includegraphics[width=\linewidth]{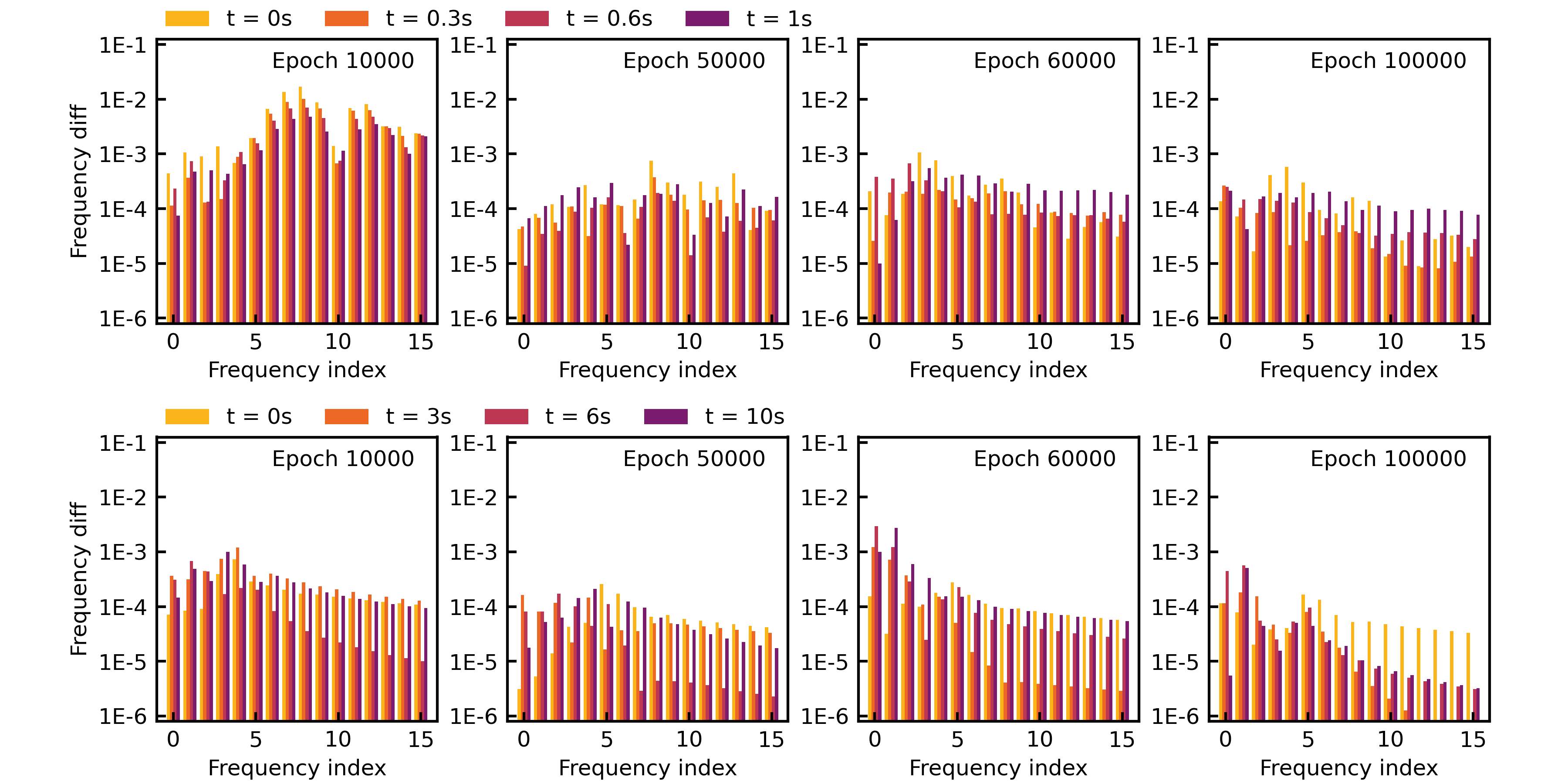}
    	\caption{Frequency error of diffusion equation (top) and wave equation (bottom).}
    	\label{fig:diffusion_wave_freq}
        \end{figure}

        From the frequency diagram, because the frequency amplitude of the function used for the diffusion equation decreases with frequency, the error at each frequency does not exhibit a significant decrease. However, it can still be observed that the fitting preference of the neural network for each frequency has changed before and after switching the loss function.

\section{Multi-stage Descent Phenomenon}\label{sec:multi-stage descent}

    \begin{figure}[htbp]
        \centering
        \includegraphics[width=\linewidth]{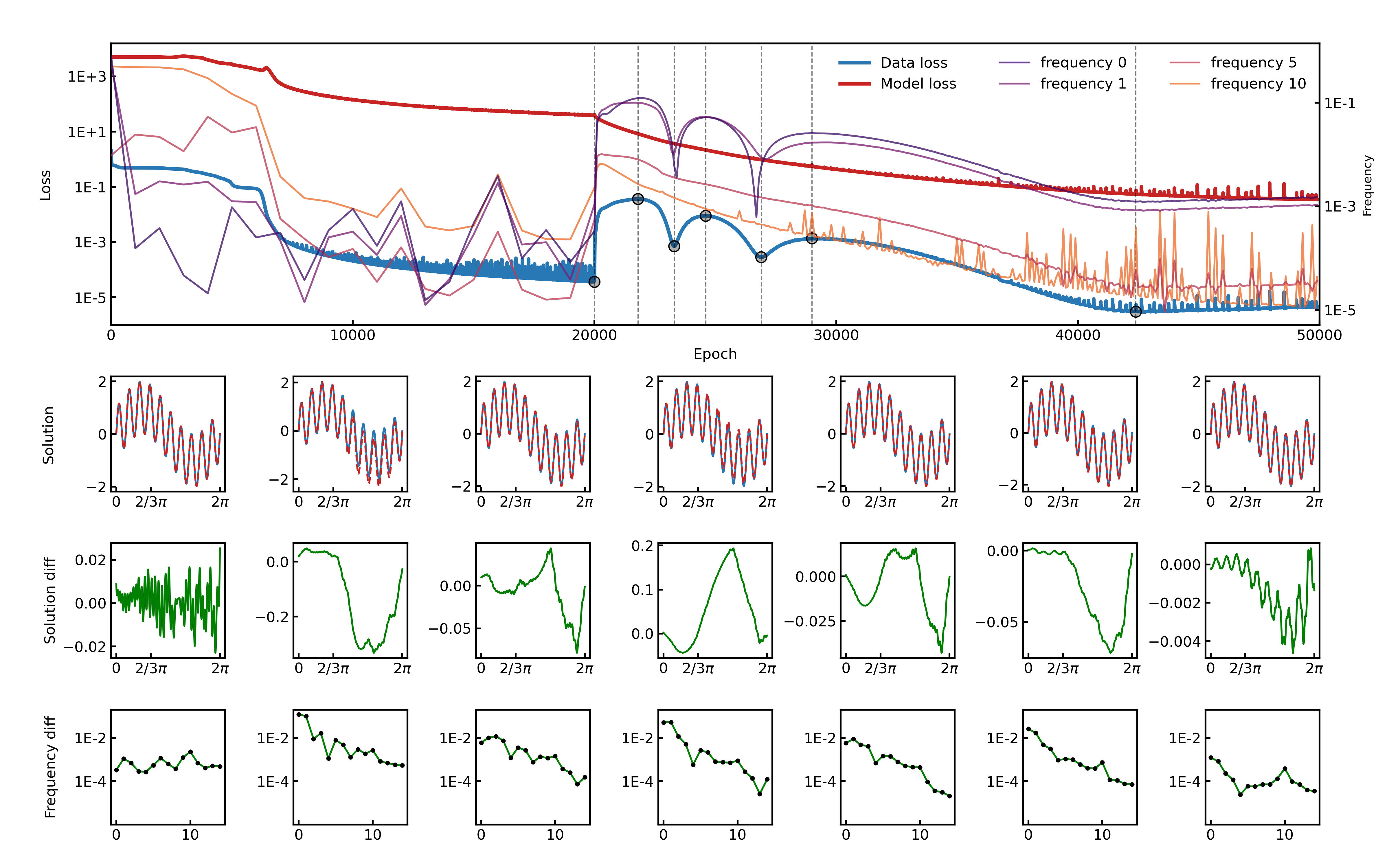}
        \caption{Multi-stage descent. We consider Eq.~\ref{eq:toy model} and use a fully connected network with 2 layers of 40 neurons in each layer for training. First, the data loss function is used to train for 20,000 epochs, then switched to the model loss function. An additional supervised learning point is added at $2/3\pi$. The weights of each part in the model loss are $(\lambda_f, \lambda_h, \lambda_g, \lambda_s)=(1,10,10,10)$. In the epoch-loss curve (top), we mark the 7 maximum or minimum points of the data loss curve. The 7 subgraphs in each subsequent row correspond to the states at these 7 points. The second row shows the predicted values of the neural network (red dotted line) and the exact solution (blue solid line). The third row is the curve of $u-u_{\vtheta}$. The fourth row is the Fourier transform of $u-u_{\vtheta}$.}
        \label{fig:multi-stage descent}
    \end{figure}
    
    We further observe the subsequent training process of the neural network for Equation \ref{eq:toy model} after switching to model loss. The difference from the previous experiment is that here we add a supervised learning point at $x=\frac{2}{3}\pi$. In Fig. \ref{fig:multi-stage descent}, we plot the training error for 80,000 epochs after switching to model loss at 20,000 epochs and the prediction errors of the neural network at different stages. Since we previously noticed that the error increase is always accompanied by an overall shift in the predicted values, we suspect that the low-frequency error cannot be well-constrained during model loss training. We plot the Fourier transform of the error in the last row of Fig. \ref{fig:multi-stage descent}. We find that during the continuous decline of model loss, data loss exhibits a multi-stage descent. When training reaches a maximum value, the error is relatively smooth, and the Fourier transform shows that the error decreases with frequency, with the low-frequency error at a relatively large level. When reaching a minimum value point, the low-frequency error decreases. However, subsequent training causes the low-frequency error to change sign, returning to another maximum value. This phenomenon demonstrates that model loss makes it difficult to constrain the low-frequency error.
    
\section{Frequency Bias for NN with model loss}\label{sec:frequency bias}
    Numerous studies have established that under the data loss setting, neural networks with common activation functions such as ReLU and tanh exhibit a frequency principle, fitting low frequencies first and then progressively capturing higher frequencies \cite{xu2019frequency,xu2019training,rahaman2019spectral,xu2022overview}. This frequency principle has been considered crucial for understanding the good generalization properties of neural networks. However, in our model loss setting, we discover that the frequency preference of neural networks differs from that observed under the data loss setting. This difference in frequency preference provides an explanation for the sudden jump in data loss when switching from data loss to model loss during training.

    To better understand the frequency preference during model loss training, we model the dynamic behavior of using model loss to train the Poisson problem. The one-dimensional Poisson problem can be described as:
    
    \begin{align}
        \Delta u(\vx) &= f(\vx), \vx\in \Omega,\\
    \end{align}

     where $u: \Omega \rightarrow \mathbb{R}$ is the unknown function to be solved, $f: \Omega \rightarrow \mathbb{R}$ is a given source term, and $\Omega \subset \mathbb{R}$ is the problem domain. The boundary conditions are not specified here, as they can be incorporated into the supervised learning points. Denoting $u_0(\vx)$ and $u(\vx, \vtheta)$ as the exact solution and the neural network approximation, respectively, the loss function can be simplified as:
    
    \begin{equation}
        R_S(\vtheta)=\frac{1}{2}\sum_{i\in S_1}(\Delta u(\vx_i,\vtheta)-\Delta u_0(\vx_i))^2+\gamma\cdot\frac{1}{2}\sum_{i\in S_2}(u(\vx_i,\vtheta)-u_0(\vx_i))^2,
    \end{equation}
    where $\gamma$ is a weight used to balance the two error terms, $S_1$ and $S_2$ are the sets of collocation points for the governing equation and the supervised learning points, respectively.

    We consider a two-layer DNN structure $u(\vx, \vtheta)=\va\sigma(\vw^T\vx+\vb)$ and use the GD algorithm to train it. Thus the parameters follow the following dynamics:

    \begin{equation}
        \begin{cases}
         \dot{\vtheta} = -\nabla_{\vtheta}R_S(\vtheta), \\
         \theta(0) = \theta_0, 
     \end{cases}
    \end{equation}
    
    For a 2-layer infinite-width neural network that conforms to the linear frequency principle, by defining $v(\vx, \vtheta)=u(\vx, \vtheta)-u_0(\vx),\ \kappa = \frac{\Gamma(d/2)}{2\sqrt{2}\pi^{(d+1)/2}\sigma_b}$ and 
    
    $$h(r^\alpha, g_i, g_j)=\Exp_{a,r}[r^\alpha\fF[g_i](\frac{\norm{\vxi}}{r})]\fF[g_j](-\frac{\norm{\vxi}}{r})],$$ 

    where the Fourier transform operator $\fF$ is defined by

    $$\fF[g](\xi)=\fF_{x\rightarrow\xi}[\xi]=\int_{\sR}g(x)e^{-2\pi i\xi x}\diff x.$$

    Here we define 5 functions $g_i(x)$ to simplify the representation of neural network derivation,

    \begin{align}
        g_1(z) &= \left(\begin{array}{c}\partial_{\va}[\va\sigma(z)]\\\partial_{\vb}[\va\sigma(z)]\end{array}\right)\nonumber\\
        g_2(z) &= a\sigma'(z)\nonumber\\
        g_3(z) &= \left(\begin{array}{c}\partial_{\va}[\va\sigma''(z)]\\\partial_{\vb}[\va\sigma''(z)]\end{array}\right)\nonumber\\
        g_4(z) &= 2a\sigma''(z)\nonumber\\
        g_5(z) &= a\sigma'''(z)\nonumber
    \end{align}
    
    we can prove the following theorem:

    \begin{theorem}[Dynamics for NN with model loss]\label{thm:LFP PINN}
        The dynamics have the following expression in the frequency domain for all $\phi\in\vS(\sR^d)$:
        \begin{align}
            \langle\partial_t\fF[\Delta v], \phi\rangle &= - \langle\fL_{\Delta}[\fF[v_\rho]],\phi\rangle,\\
            \langle\partial_t\fF[v], \phi\rangle &= - \langle\fL[\fF[v_\rho]],\phi\rangle,
        \end{align}

        Where $v_\rho(x) = v(x)\rho(x) = v(x)\rho_1(x)+v(x)\rho_2(x)$ with empirical density $\rho_j(x)=\sum_{i\in S_j} \delta(x-x_i)$ and 
        
        \begin{align}
            \fL_{\Delta}[\fF[v_\rho]] &= \frac{\kappa}{\norm{\vxi}^{d-1}}h(r^3,g_3,g_3)\fF[v''_{\rho_1}](\vxi)+\frac{\kappa}{\norm{\vxi}^{d-1}}h(r,g_4,g_4)\fF[v''_{\rho_1}](\vxi)\nonumber\\
            &+\nabla\cdot(\frac{\kappa}{\norm{\vxi}^{d-1}}h(r^2,g_5,g_4))\fF[v''_{\rho_1}](\vxi)-\frac{\kappa}{\norm{\vxi}^{d-1}}h(r^2,g_4,g_5)\nabla\fF[v''_{\rho_1}](\vxi)\nonumber\\
            &-\nabla\cdot(\frac{\kappa}{\norm{\vxi}^{d-1}}h(r^3,g_5,g_5)\nabla\fF[v''_{\rho_1}](\vxi))\nonumber\\
            &+ \gamma\frac{\kappa}{\norm{\vxi}^{d-1}}h(r,g_3,g_1)\fF[v_{\rho_2}](\vxi)
            -\gamma\frac{\kappa}{\norm{\vxi}^{d-1}}h(1,g_4,g_2)\nabla\fF[v_{\rho_2}](\vxi)\nonumber\\
            &-\gamma\nabla\cdot(\frac{\kappa}{\norm{\vxi}^{d-1}}h(r^2,g_5,g_2)\nabla\fF[v_{\rho_2}](\vxi)),\\
            \fL[\fF[v_\rho]] &= \frac{\kappa}{\norm{\vxi}^{d-1}}h(r,g_1,g_3)\fF[v''_{\rho_1}](\vxi)
            -\frac{\kappa}{\norm{\vxi}^{d-1}}h(1,g_2,g_4)\nabla\fF[v''_{\rho_1}](\vxi)\nonumber\\
            &-\nabla\cdot(\frac{\kappa}{\norm{\vxi}^{d-1}}h(r^2,g_2,g_5)\nabla\fF[v''_{\rho_1}](\vxi))\nonumber\\
            &+\gamma\frac{\kappa}{\norm{\vxi}^{d-1}}h(\frac{1}{r},g_1,g_1)\fF[v_{\rho_2}](\vxi)
            -\gamma\nabla\cdot(\frac{\kappa}{\norm{\vxi}^{d-1}}h(\frac{1}{r},g_2,g_2)\nabla\fF[v_{\rho_2}](\vxi))
        \end{align}
    \end{theorem}

    For the $\tanh$ activation function, we have
    \begin{align}
        \fF[g_1](\xi) &= \left(\begin{array}{c}-i\pi\csch(\pi^2\xi)\\2\pi^2a\xi\csch(\pi^2\xi)\end{array}\right)\nonumber\\
        \fF[g_2](\xi) &= 2\pi^2a\xi\csch(\pi^2\xi)\nonumber\\
        \fF[g_3](\xi) &= \left(\begin{array}{c}4\pi^3i\xi^2\csch(\pi^2\xi)\\-8\pi^4a\xi^3\csch(\pi^2\xi)\end{array}\right)\nonumber\\
        \fF[g_4](\xi) &= 8\pi^3ia\xi^2\csch(\pi^2\xi)\nonumber\\
        \fF[g_5](\xi) &= -8\pi^4a\xi^3\csch(\pi^2\xi)\nonumber
    \end{align}
    
    If we temporarily ignore all derivative terms, we can get
    \begin{align}
        \fL_{\Delta}[\fF[v_\rho]] 
        &=\frac{\kappa}{\norm{\vxi}^{d-1}}\Exp_{a,r}[(\frac{64\pi^8a^2\norm{\vxi}^6}{r^3}+\frac{16\pi^6\norm{\vxi}^4}{r}+\frac{64\pi^6a^2\norm{\vxi}^4}{r^5})]\csch^2(\frac{\pi^2\norm{\vxi}}{r})\fF[v''_{\rho_1}](\vxi)\nonumber\\
        &+ \gamma\frac{\kappa}{\norm{\vxi}^{d-1}}\Exp_{a,r}[(-\frac{16\pi^6a^2\norm{\vxi}^4}{r^3}-\frac{4\pi^4\norm{\vxi}^2}{r})\csch^2(\frac{\pi^2\norm{\vxi}}{r})]\fF[v_{\rho_2}](\vxi)
    \end{align}

    \begin{align}
        \fL[\fF[v_\rho]] 
        &=\frac{\kappa}{\norm{\vxi}^{d-1}}\Exp_{a,r}[(\frac{64\pi^8a^2\norm{\vxi}^6}{r^3}+\frac{16\pi^6\norm{\vxi}^4}{r})\csch^2(\frac{\pi^2\norm{\vxi}}{r})]\fF[v_{\rho_1}](\vxi)\nonumber\\
        &+\gamma\frac{\kappa}{\norm{\vxi}^{d-1}}\Exp_{a,r}[(\frac{4\pi^4a^2\norm{\vxi}^2}{r^3}+\frac{\pi^2}{r})\csch^2(\frac{\pi^2\norm{\vxi}}{r})]\fF[v_{\rho_2}](\vxi)
    \end{align}

\begin{figure}[htbp]
	\centering
	\includegraphics[width=0.7\linewidth]{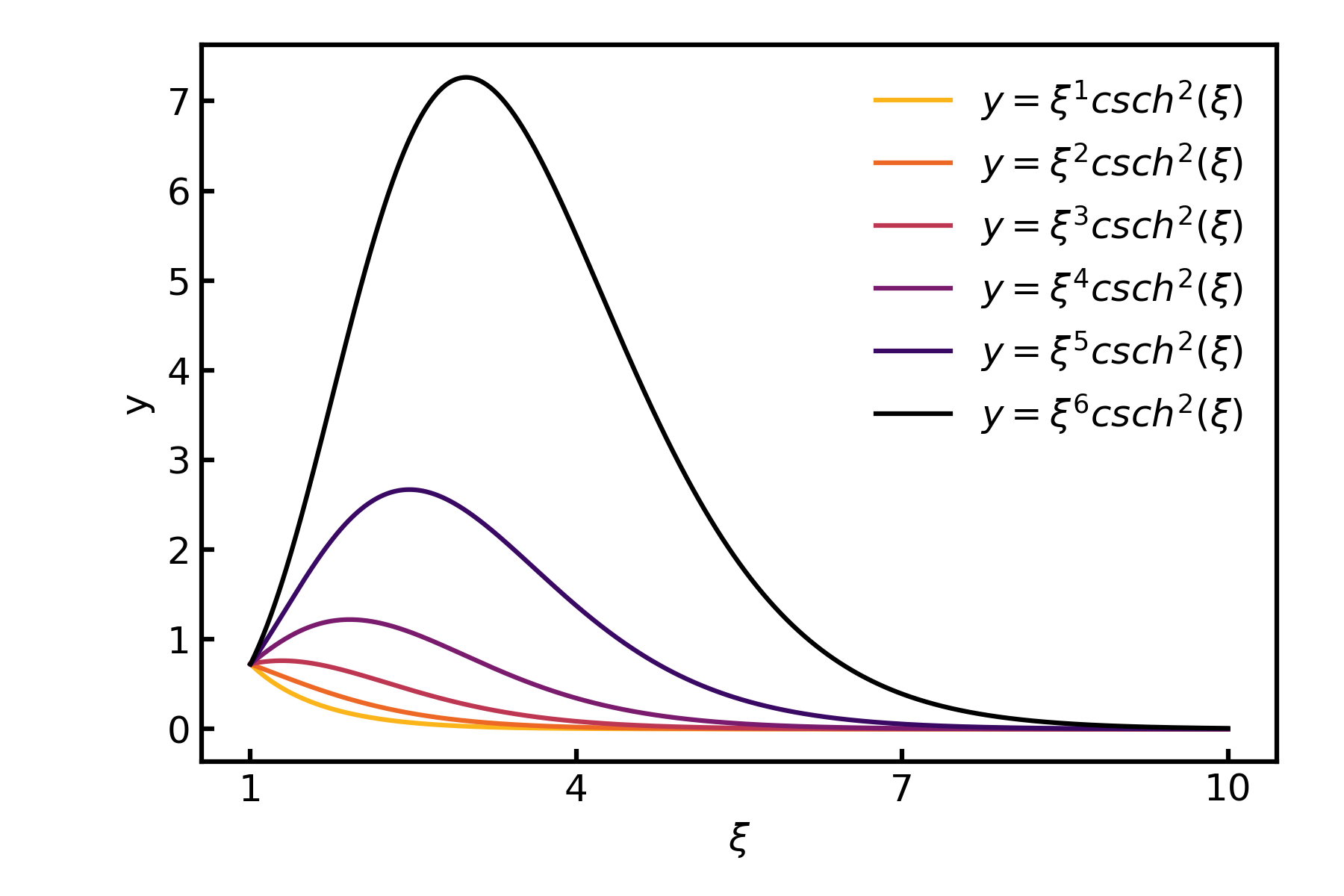}
	\caption{Values of $\xi^n\csch^2(\xi)$ for different polynomial degrees $n$.}
	\label{fig:xn_cschx}
    \end{figure}

    For one-dimensional problems, the convergence rate of frequency is governed by the function $\xi^n\csch^2(\xi)$. As shown in Fig. \ref{fig:xn_cschx}, when the polynomial degree $n$ is large, $\xi^n\csch^2(\xi)$ increases with frequency within a certain range. Consequently, high frequencies exhibit faster convergence rates in this regime. However, once the frequency surpasses a certain threshold, $\xi^n\csch^2(\xi)$ rapidly decays with increasing frequency. From the perspective of a broader frequency spectrum, this behavior suggests that model loss tends to prioritize the learning of low and medium frequencies.
    
    These findings highlight the complex interplay between frequency components and the convergence dynamics of neural networks under the model loss setting. The preferential learning of low and medium frequencies by model loss may have implications for the overall performance and generalization ability of the trained networks. Further research is needed to fully understand the impact of this frequency bias on the effectiveness of model loss-based training approaches for solving partial differential equations and other related problems.

\section{Discussion}

    The loss jump phenomenon observed when switching from data loss to model loss highlights the complex interplay between loss functions, frequency bias, and convergence behavior in neural networks for solving partial differential equations. The sudden increase in data loss suggests that the solutions obtained using data loss may not provide a suitable starting point for model loss optimization, challenging the conventional wisdom of pre-training with data loss.
    
    The multi-stage descent phenomenon suggests that model loss imposes weak constraints on low-frequency errors, which could have implications for the accuracy and reliability of the trained networks. Furthermore, our theoretical analysis reveals that model loss exhibits a frequency bias that differs from the well-established frequency principle observed in networks trained with data loss. Within a certain frequency range, high frequencies converge faster under model loss, while low and medium frequencies are prioritized when considering the entire spectrum. This difference in frequency preference provides a plausible explanation for the loss jump phenomenon.
    
    Future research directions could include the development of adaptive training strategies, frequency-dependent weighting schemes, and regularization techniques to mitigate the impact of the loss jump and improve the performance of model loss-based approaches in scientific computing and engineering applications.

\bibliographystyle{unsrt}  
\bibliography{library}  






\begin{appendix}
    \section{Proof of Theorem \ref{thm:LFP PINN}}
        Consider the following gradient descent dynamics of the empirical risk $R_S$ of a network function $f(\cdot, \vtheta)$ parameterized by $\vtheta$:
        \begin{equation}
            \begin{cases}
             \dot{\vtheta} = -\nabla_{\vtheta}R_S(\vtheta), \\
             \theta(0) = \theta_0, 
         \end{cases}
        \end{equation}
        where
        \begin{equation}
            R_S(\vtheta)=\frac{1}{2}\sum_{i\in S_1}(u''(\vx_i,\vtheta)-u''(x_i))^2+\gamma\cdot\frac{1}{2}\sum_{i\in S_2}(u(\vx_i,\vtheta)-u(\vx_i))^2
        \end{equation}
    
        Thus the training dynamics of residual function $v''(\cdot, \vtheta) = u''(\cdot,\vtheta)-u''(\cdot)$ is
        \begin{align}
            \frac{\diff}{\diff t}v''(\vx, \vtheta) &= \nabla_{\vtheta}v''(\vx,\vtheta)\dot{\vtheta} = 
            -\nabla_{\vtheta}v''(\vx,\vtheta)\cdot\nabla_{\vtheta}R_S(\vtheta)\\
            &=-\sum_{i\in S_1}\nabla_{\vtheta}v''(\vx,\vtheta)\nabla_{\vtheta}v''(\vx_i,\vtheta)v''(\vx_i,\vtheta)\\
            &-\gamma \sum_{j\in S_2}\nabla_{\vtheta}v''(\vx,\vtheta)\nabla_{\vtheta}v(\vx_j,\vtheta)v(\vx_j,\vtheta)\\
            &=-\sum_{i\in S_1}K_1(\vx,\vx_i)v''(\vx_i,\vtheta) - \gamma \sum_{j\in S_2}K_2(\vx,\vx_j)v(\vx_j,\vtheta)
        \end{align}
    
        where 
        \begin{align}
            K_1(\vx,\vx_i) &= \nabla_{\vtheta}v''(\vx,\vtheta)\nabla_{\vtheta}v''(\vx_i,\vtheta)\\
            K_2(\vx,\vx_j) &= \nabla_{\vtheta}v''(\vx,\vtheta)\nabla_{\vtheta}v(\vx_j,\vtheta)
        \end{align}

        Let $\rho_1(\vx) = \sum_{i\in S_1}\delta(x-x_i), \rho_2(\vx) = \sum_{j\in S_2}\delta(x-x_j)$, then we have
    
        \begin{equation}
            \frac{\diff}{\diff t}v''(x, \vtheta) = 
            -\int_{\sR^d}K_1(\vx,\vx')(t)v_{\rho_1}''(\vx',t)\diff\vx' 
            - \gamma \int_{\sR^d}K_2(\vx,\vx'')(t)v_{\rho_2}''(\vx'',t)\diff\vx''
        \end{equation}

        And the training dynamics of residual function $v(\cdot, \vtheta) = u(\cdot,\vtheta)-u(\cdot)$ is
        \begin{align}
            \frac{\diff}{\diff t}v(\vx, \vtheta) &= \nabla_{\vtheta}v(\vx,\vtheta)\dot{\vtheta} = 
            -\nabla_{\vtheta}v(\vx,\vtheta)\cdot\nabla_{\vtheta}R_S(\vtheta)\\
            &=-\sum_{i\in S_1}\nabla_{\vtheta}v(\vx,\vtheta)\nabla_{\vtheta}v''(\vx_i,\vtheta)v''(\vx_i,\vtheta)\\
            &-\gamma \sum_{j\in S_2}\nabla_{\vtheta}v(\vx,\vtheta)\nabla_{\vtheta}v(\vx_j,\vtheta)v(\vx_j,\vtheta)\\
            &=-\sum_{i\in S_1}K_3(\vx,\vx_i)v''(\vx_i,\vtheta) - \gamma \sum_{j\in S_2}K_4(\vx,\vx_j)v(\vx_j,\vtheta)
        \end{align}
    
        where 
        \begin{align}
            K_3(\vx,\vx_i) &= \nabla_{\vtheta}v(\vx,\vtheta)\nabla_{\vtheta}v''(\vx_i,\vtheta)\\
            K_4(\vx,\vx_j) &= \nabla_{\vtheta}v(\vx,\vtheta)\nabla_{\vtheta}v(\vx_j,\vtheta)
        \end{align}
    
        Therefore,
    
        \begin{equation}
            \frac{\diff}{\diff t}v(x, \vtheta) = 
            -\int_{\sR^d}K_3(\vx,\vx')(t)v_{\rho_1}''(\vx',t)\diff\vx' 
            - \gamma \int_{\sR^d}K_4(\vx,\vx'')(t)v_{\rho_2}(\vx'',t)\diff\vx''
        \end{equation}
    
        By mean field assumption,
    
        \begin{equation}
            \frac{\diff}{\diff t}v''(x, \vtheta) = 
            -\int_{\sR^d}\overline{K_1}(\vx,\vx')v_{\rho_1}''(\vx',t)\diff\vx' 
            - \gamma \int_{\sR^d}\overline{K_2}(\vx,\vx'')v_{\rho_2}(\vx'',t)\diff\vx''
        \end{equation}
    
        \begin{equation}
            \frac{\diff}{\diff t}v(x, \vtheta) = 
            -\int_{\sR^d}\overline{K_3}(\vx,\vx')v_{\rho_1}''(\vx',t)\diff\vx' 
            - \gamma \int_{\sR^d}\overline{K_4}(\vx,\vx'')v_{\rho_2}(\vx'',t)\diff\vx''
        \end{equation}
    
        and 
        \begin{align}
            \overline{K_1}(\vx,\vx') &= \Exp_{\vq}[\nabla_{\vq}\sigma^*{''}(\vx,\vq)\cdot\nabla_{\vq}\sigma^*{''}(\vx',\vq)]\\
            \overline{K_2}(\vx,\vx') &= \Exp_{\vq}[\nabla_{\vq}\sigma^*{''}(\vx,\vq)\cdot\nabla_{\vq}\sigma^*(\vx',\vq)]\\
            \overline{K_3}(\vx,\vx') &= \Exp_{\vq}[\nabla_{\vq}\sigma^*(\vx,\vq)\cdot\nabla_{\vq}\sigma^*{''}(\vx',\vq)]\\
            \overline{K_4}(\vx,\vx') &= \Exp_{\vq}[\nabla_{\vq}\sigma^*(\vx,\vq)\cdot\nabla_{\vq}\sigma^*(\vx',\vq)]
        \end{align}

        where $\sigma^*(\vx',\vq) = \va\sigma(\vw^T\vx+\vb)$.

        \begin{lemma}[Dynamics for $v''$]\label{lem.dynamic_for_uxx}
            The dynamics has the following expression in the frequency domain for all $\phi\in\vS(\sR^d)$:
            \begin{equation}
                \langle\partial_t\fF[v''], \phi\rangle = - \langle\fL_{\Delta}[\fF[v_\rho]],\phi\rangle,
            \end{equation}
            where $\fL_{\Delta}[\cdot]$ is given by
            \begin{equation}
                \fL_{\Delta}[\fF[v_\rho]] = \int_{\sR^d}\hat{K_1}(\xi, \xi')\fF[v_{\rho_1}''](\xi')\diff\xi' + \gamma\int_{\sR^d}\hat{K_2}(\xi, \xi'')\fF[v_{\rho_2}](\xi'')\diff\xi''
            \end{equation}
            and 
            \begin{equation}
                \hat{K}_1(\vxi, \vxi') = \Exp_{\vq}[\fF_{\vx\to\vxi}[\nabla_{\vq}\sigma^*{''}(\vx,\vq)]
                \overline{\fF_{\vx'\to\vxi'}[\nabla_{\vq}\sigma^*{''}(\vx',\vq)]}] \triangleq \Exp_{\vq}[\hat{K}_{1,q}(\vxi, \vxi')]
            \end{equation}
            \begin{equation}
                \hat{K}_2(\vxi, \vxi'') = \Exp_{\vq}[\fF_{\vx\to\vxi}[\nabla_{\vq}\sigma^*{''}(\vx,\vq)]
                \overline{\fF_{\vx''\to\vxi''}[\nabla_{\vq}\sigma^*(\vx'',\vq)]}] \triangleq \Exp_{\vq}[\hat{K}_{2,q}(\vxi, \vxi')]
            \end{equation}
        \end{lemma}
    
        Define $\kappa = \frac{\Gamma(d/2)}{2\sqrt{2}\pi^{(d+1)/2}\sigma_b}$ and $h(r^\alpha, g_i, g_j)=\Exp_{a,r}[r^\alpha\fF[g_i](\frac{\norm{\vxi}}{r})]\fF[g_j](-\frac{\norm{\vxi}}{r})]$
    
        \begin{proof}
            For any $\phi \in \fS\left(\sR^d\right)$. since $\partial_t v''$ is in $\fS'\left(\sR^d\right)$ and locally integrable, we have
       
            \begin{align}
                \left\langle\partial_t \fF[v''], \phi\right\rangle & =\left\langle\partial_t v'', \fF[\phi]\right\rangle \nonumber\\
                & =\int_{\sR^d} \partial_t v''(\vx, t) \int_{\sR^d} \phi(\vxi) \mathrm{e}^{-\mathrm{i} 2 \pi \vx \cdot \vxi} \diff \vxi \diff \vx \nonumber\\
                & =-\int_{\sR^d} [\int_{\sR^d} \overline{K_1}\left(\vx, \vx'\right) v_{\rho_1}''\left(\vx'\right) \diff \vx'+\int_{\sR^d} \overline{K_2}\left(\vx, \vx''\right) v_{\rho_2}\left(\vx''\right) \diff \vx''] \int_{\sR^d} \phi(\vxi) \mathrm{e}^{-\mathrm{i} 2 \pi \vx \cdot \vxi} \diff \vxi \diff \vx \nonumber\\
                & =-\int_{\sR^{3 d}} \overline{K_1}\left(\vx, \vx'\right) v_{\rho_1}''\left(\vx'\right) \diff \vx' \phi(\vxi) \mathrm{e}^{-\mathrm{i} 2 \pi \vx \cdot \vxi} \diff \vxi \diff \vx \nonumber\\
                &\quad -\int_{\sR^{3 d}} \overline{K_2}\left(\vx, \vx''\right) v_{\rho_2}\left(\vx''\right) \diff \vx'' \phi(\vxi) \mathrm{e}^{-\mathrm{i} 2 \pi \vx \cdot \vxi} \diff \vxi \diff \vx \nonumber\\
                & =-\int_{\sR^{3 d}} \Exp_{\vq}[\nabla_{\vq} \sigma^*{''}(\vx, \vq) \cdot \nabla_{\vq} \sigma^*{''}\left(\vx', \vq\right)] v_{\rho_1}''\left(\vx'\right) \diff \vx' \phi(\vxi) \mathrm{e}^{-\mathrm{i} 2 \pi x \cdot \vxi} \diff \vxi \diff \vx \nonumber\\
                &\quad-\int_{\sR^{3 d}} \Exp_{\vq}[\nabla_{\vq} \sigma^*{''}(\vx, \vq) \cdot \nabla_{\vq} \sigma^*\left(\vx'', \vq\right)] v_{\rho_2}\left(\vx''\right) \diff \vx'' \phi(\vxi) \mathrm{e}^{-\mathrm{i} 2 \pi x \cdot \vxi} \diff \vxi \diff \vx \nonumber\\
                & =-\Exp_{\vq} \int_{\sR^d} \nabla_{\vq} \sigma^*{''}\left(\vx', \vq\right) v_{\rho_1}''\left(\vx'\right) \diff \vx' \cdot \int_{\sR^{2 d}} \nabla_{\vq} \sigma^*{''}(\vx, \vq) \mathrm{e}^{-\mathrm{i} 2 \pi \vx \cdot \vxi} \phi(\vxi) \diff \vxi \diff \vx \nonumber\\
                &\quad-\Exp_{\vq} \int_{\sR^d} \nabla_{\vq} \sigma^*\left(\vx'', \vq\right) v_{\rho_2}\left(\vx''\right) \diff \vx'' \cdot \int_{\sR^{2 d}} \nabla_{\vq} \sigma^*{''}(\vx, \vq) \mathrm{e}^{-\mathrm{i} 2 \pi \vx \cdot \vxi} \phi(\vxi) \diff \vxi \diff \vx \nonumber\\
                & =-\Exp_{\vq} \int_{\sR^d} \nabla_{\vq} \sigma^*{''}\left(\vx', \vq\right) v_{\rho_1}''\left(\vx'\right) \diff \vx' \cdot\left\langle\fF_{\vx \rightarrow \cdot}\left[\nabla_{\vq} \sigma^*{''}(\vx, \vq)\right](\cdot), \phi(\cdot)\right\rangle \nonumber\\
                &\quad -\Exp_{\vq} \int_{\sR^d} \nabla_{\vq} \sigma^*\left(\vx'', \vq\right) v_{\rho_2}\left(\vx''\right) \diff \vx'' \cdot\left\langle\fF_{\vx \rightarrow \cdot}\left[\nabla_{\vq} \sigma^*{''}(\vx, \vq)\right](\cdot), \phi(\cdot)\right\rangle.
            \end{align}
            
            Since
            \begin{align}
                \int_{\sR^d} \nabla_{\vq} \sigma^*{''}\left(\vx', \vq\right) v_{\rho_1}''\left(\vx'\right) \diff \vx' &= \int_{\sR^d} \overline{\fF_{\vx' \rightarrow \vxi'}\left[\nabla_{\vq} \sigma^*{''}\left(\vx', \vq\right)\right]\left(\vxi'\right)} \fF_{\vx' \rightarrow \vxi'}\left[v_{\rho_1}''\right]\left(\vxi'\right) \diff \vxi'\nonumber\\
                \int_{\sR^d} \nabla_{\vq} \sigma^*\left(\vx'', \vq\right) v_{\rho_2}\left(\vx''\right) \diff \vx'' &= \int_{\sR^d} \overline{\fF_{\vx'' \rightarrow \vxi''}\left[\nabla_{\vq} \sigma^*\left(\vx'', \vq\right)\right]\left(\vxi''\right)} \fF_{\vx'' \rightarrow \vxi''}\left[v_{\rho_2}\right]\left(\vxi''\right) \diff \vxi''\nonumber,
            \end{align}
            
            we have
            \begin{align}
                \left\langle\partial_t \fF[v''], \phi\right\rangle & =-\Exp_{\vq} \int_{\sR^d} \overline{\fF_{\vx' \rightarrow \vxi'}\left[\nabla_{\vq} \sigma^*{''}\left(\vx', \vq\right)\right]\left(\vxi'\right)} \fF_{\vx' \rightarrow \vxi'}\left[v_{\rho_1}''\right]\left(\vxi'\right) \diff \vxi'\cdot\left\langle\fF_{\vx \rightarrow \cdot}\left[\nabla_{\vq} \sigma^*{''}(\vx, \vq)\right](\cdot), \phi(\cdot)\right\rangle \nonumber\\
                &\quad- \Exp_{\vq} \int_{\sR^d} \overline{\fF_{\vx'' \rightarrow \vxi''}\left[\nabla_{\vq} \sigma^*\left(\vx'', \vq\right)\right]\left(\vxi''\right)} \fF_{\vx'' \rightarrow \vxi''}\left[v_{\rho_2}\right]\left(\vxi''\right) \diff \vxi''\cdot\left\langle\fF_{\vx \rightarrow \cdot}\left[\nabla_{\vq} \sigma^*{''}(\vx, \vq)\right](\cdot), \phi(\cdot)\right\rangle \nonumber\\
                & =-\Exp_{\vq} \int_{\sR^{2 d}} \overline{\fF_{\vx' \rightarrow \vxi'}\left[\nabla_{\vq} \sigma^*{''}\left(\vx', \vq\right)\right]\left(\vxi'\right)} \cdot \fF_{\vx \rightarrow \vxi}\left[\nabla_{\vq} \sigma^*{''}(\vx, \vq)\right](\vxi) \fF_{\vx' \rightarrow \vxi'}\left[v_{\rho_1}''\right]\left(\vxi'\right) \diff \vxi' \phi(\vxi) \diff \vxi \nonumber\\
                &\quad -\Exp_{\vq} \int_{\sR^{2 d}} \overline{\fF_{\vx'' \rightarrow \vxi''}\left[\nabla_{\vq} \sigma^*\left(\vx'', \vq\right)\right]\left(\vxi''\right)} \cdot \fF_{\vx \rightarrow \vxi}\left[\nabla_{\vq} \sigma^*{''}(\vx, \vq)\right](\vxi) \fF_{\vx'' \rightarrow \vxi''}\left[v_{\rho_2}\right]\left(\vxi''\right) \diff \vxi'' \phi(\vxi) \diff \vxi \nonumber\\
                & =-\int_{\sR^{2 d}} \hat{K}_1\left(\vxi, \vxi'\right) \fF\left[v_{\rho_1}''\right]\left(\vxi'\right) \diff \vxi' \phi(\vxi) \diff \vxi
                --\int_{\sR^{2 d}} \hat{K}_2\left(\vxi, \vxi''\right) \fF\left[v_{\rho_2}\right]\left(\vxi''\right) \diff \vxi'' \phi(\vxi) \diff \vxi\nonumber\\
                & =-\left\langle\fL_{\Delta}\left[\fF\left[v_\rho\right]\right], \phi\right\rangle .
            \end{align}
            
        \end{proof}

        In the same way as the proof of Lemma \ref{lem.dynamic_for_uxx}, we can get
        
        \begin{lemma}[Dynamics for $v$]\label{lem.dynamic_for_u}
            The dynamics has the following expression in the frequency domain for all $\phi\in\vS(\sR^d)$:
            \begin{equation}
                \langle\partial_t\fF[v], \phi\rangle = - \langle\fL[\fF[v_\rho]],\phi\rangle,
            \end{equation}
            where $\fL[\cdot]$ is given by
            \begin{equation}
                \fL[\fF[v_\rho]] = \int_{\sR^d}\hat{K_3}(\xi, \xi')\fF[v_{\rho_1}''](\xi')\diff\xi' + \gamma\int_{\sR^d}\hat{K_4}(\xi, \xi'')\fF[v_{\rho_2}](\xi'')\diff\xi''
            \end{equation}
            and 
            \begin{equation}
                \hat{K}_3(\vxi, \vxi') = \Exp_{\vq}[\fF_{\vx\to\vxi}[\nabla_{\vq}\sigma^*(\vx,\vq)]
                \overline{\fF_{\vx'\to\vxi'}[\nabla_{\vq}\sigma^*{''}(\vx',\vq)]}] \triangleq \Exp_{\vq}[\hat{K}_{3,q}(\vxi, \vxi')]
            \end{equation}
            \begin{equation}
                \hat{K}_4(\vxi, \vxi'') = \Exp_{\vq}[\fF_{\vx\to\vxi}[\nabla_{\vq}\sigma^*(\vx,\vq)]
                \overline{\fF_{\vx''\to\vxi''}[\nabla_{\vq}\sigma^*(\vx'',\vq)]}] \triangleq \Exp_{\vq}[\hat{K}_{4,q}(\vxi, \vxi')]
            \end{equation}
        \end{lemma}

    The kernel function can be divided into 2 parts $\hat{K}_{1,q}(\vxi, \vxi') = \hat{K}_{1,\va b}(\vxi, \vxi') + \hat{K}_{1,\vw}(\vxi, \vxi')$ and $\hat{K}_{2,q}(\vxi, \vxi') = \hat{K}_{2,\va b}(\vxi, \vxi') + \hat{K}_{2,\vw}(\vxi, \vxi')$, where
    \begin{align}
        \hat{K}_{1,\va b}(\vxi, \vxi') &= \fF_{\vx\to\vxi}[\vw^T\vw g_3(\vw^T\vx+b)]
                \overline{\fF_{\vx'\to\vxi'}[\vw^T\vw g_3(\vw^T\vx'+b)]}\\
        \hat{K}_{1,\vw}(\vxi, \vxi') &= \fF_{\vx\to\vxi}[\vw g_4(\vw^T\vx+b)+\vw^T\vw \vx g_5(\vw^T\vx+b)]
                \overline{\fF_{\vx'\to\vxi'}[\vw g_4(\vw^T\vx'+b)+\vw^T\vw \vx' g_5(\vw^T\vx'+b)]}\\
        \hat{K}_{2,\va b}(\vxi, \vxi') &= \fF_{\vx\to\vxi}[\vw^T\vw g_3(\vw^T\vx+b)]
                \overline{\fF_{\vx'\to\vxi'}[g_1(\vw^T\vx'+b)]}\\
        \hat{K}_{2,\vw}(\vxi, \vxi') &= \fF_{\vx\to\vxi}[\vw g_4(\vw^T\vx+b)+\vw^T\vw \vx g_5(\vw^T\vx+b)]
                \overline{\fF_{\vx'\to\vxi'}[\vx g_2(\vw^T\vx'+b)]} 
    \end{align}

    With the method proposed in \cite{luo2022exact}, we have:
    \begin{align}
        \Exp_{\vw, b}[\langle\hat{K}_{1,\va b}, \phi\times\psi\rangle] &= \Exp[\frac{1}{\sqrt{2\pi}\sigma_b}\int_\sR(\vw^T\vw)^2\phi(\eta\vw)\psi(\eta\vw)\fF[g_3](\eta)\overline{\fF[g_3](\eta)}\diff \eta] \\
        &=\frac{1}{\sqrt{2\pi}\sigma_b}\int_{\sR^{d+1}} (\vw^T\vw)^2\phi(\eta\vw)\psi(\eta\vw)\fF[g_3](\eta)\overline{\fF[g_3](\eta)}\rho_{\vw}(\vw)\diff \vw \diff\eta \\
        &=\frac{\Gamma(d/2)}{2\sqrt{2}\pi^{\frac{d+1}{2}}\sigma_b}\int_{\sR^d}\phi(\vxi)\psi(\vxi) \int_{\sR^+}\frac{r^4}{r\norm{\vxi}^{d-1}}\fF[g_3](\frac{\norm{\vxi}}{r})\overline{\fF[g_3](\frac{\norm{\vxi}}{r})}\rho_{r}(r)\diff r \diff \vxi
    \end{align}

    Let $\psi=\fF[v_\rho'']$, we have,

    \begin{equation*}
        \fL_{1,\va b}[\fF[v_{\rho_1}'']](\vxi)=\frac{\Gamma(d/2)}{2\sqrt{2}\pi^{\frac{d+1}{2}}\sigma_b\norm{\vxi}^{d-1}}
        \Exp_{a,r}[r^3\fF[g_3](\frac{\norm{\vxi}}{r})\fF[g_3](-\frac{\norm{\vxi}}{r})]\fF[v_{\rho_1}''](\vxi)
    \end{equation*}

    In the same way, we can get
    \begin{align}
        \fL_{1,\vw}[\fF[v_{\rho_1}'']](\vxi)&=\frac{\Gamma(d/2)}{2\sqrt{2}\pi^{\frac{d+1}{2}}\sigma_b\norm{\vxi}^{d-1}}
        \Exp_{a,r}[r\fF[g_4](\frac{\norm{\vxi}}{r})\fF[g_4](-\frac{\norm{\vxi}}{r})]\fF[v_{\rho_1}''](\vxi)\nonumber\\
        &+\frac{\Gamma(d/2)}{2\sqrt{2}\pi^{\frac{d+1}{2}}\sigma_b}\nabla\cdot (\Exp_{a,r}[\frac{r^2}{\norm{\vxi}^{d-1}}\fF[g_5](\frac{\norm{\vxi}}{r})\fF[g_4](-\frac{\norm{\vxi}}{r})])\fF[v_{\rho_1}''](\vxi)\nonumber\\
        &-\frac{\Gamma(d/2)}{2\sqrt{2}\pi^{\frac{d+1}{2}}\sigma_b}\Exp_{a,r}[\frac{r^2}{\norm{\vxi}^{d-1}}\fF[g_4](\frac{\norm{\vxi}}{r})\fF[g_5](-\frac{\norm{\vxi}}{r})]\nabla\fF[v_{\rho_1}''](\vxi)\nonumber\\
        &-\frac{\Gamma(d/2)}{2\sqrt{2}\pi^{\frac{d+1}{2}}\sigma_b}\nabla\cdot(\Exp_{a,r}[\frac{r^3}{\norm{\vxi}^{d-1}}\fF[g_5](\frac{\norm{\vxi}}{r})\fF[g_5](-\frac{\norm{\vxi}}{r})]\nabla\fF[v_{\rho_1}''](\vxi))\nonumber
    \end{align}
    \begin{equation*}
        \fL_{2,\va b}[\fF[v_{\rho_2}]](\vxi)=\frac{\Gamma(d/2)}{2\sqrt{2}\pi^{\frac{d+1}{2}}\sigma_b\norm{\vxi}^{d-1}}
        \Exp_{a,r}[r^3\fF[g_3](\frac{\norm{\vxi}}{r})\fF[g_1](-\frac{\norm{\vxi}}{r})]\fF[v_{\rho_2}](\vxi)
    \end{equation*}
    \begin{align}
        \fL_{2,\vw}[\fF[v_{\rho_2}]](\vxi)&=\frac{\Gamma(d/2)}{2\sqrt{2}\pi^{\frac{d+1}{2}}\sigma_b}\Exp_{a,r}[\frac{1}{\norm{\vxi}^{d-1}}\fF[g_4](\frac{\norm{\vxi}}{r})\fF[g_2](-\frac{\norm{\vxi}}{r})]\nabla\fF[v_{\rho_2}](\vxi)\nonumber\\
        &-\frac{\Gamma(d/2)}{2\sqrt{2}\pi^{\frac{d+1}{2}}\sigma_b}\nabla\cdot(\Exp_{a,r}[\frac{r^2}{\norm{\vxi}^{d-1}}\fF[g_5](\frac{\norm{\vxi}}{r})\fF[g_2](-\frac{\norm{\vxi}}{r})]\nabla\fF[v_{\rho_2}](\vxi))\nonumber
    \end{align}
    \begin{equation*}
        \fL_{3,\va b}[\fF[v_{\rho_1}'']](\vxi)=\frac{\Gamma(d/2)}{2\sqrt{2}\pi^{\frac{d+1}{2}}\sigma_b\norm{\vxi}^{d-1}}
        \Exp_{a,r}[r\fF[g_1](\frac{\norm{\vxi}}{r})\fF[g_3](-\frac{\norm{\vxi}}{r})]\fF[v_{\rho_1}''](\vxi)
    \end{equation*}
    \begin{align*}
        \fL_{3,\vw}[\fF[v_{\rho_1}'']](\vxi)&=\frac{\Gamma(d/2)}{2\sqrt{2}\pi^{\frac{d+1}{2}}\sigma_b}\Exp_{a,r}[\frac{1}{\norm{\vxi}^{d-1}}\fF[g_2](\frac{\norm{\vxi}}{r})\fF[g_4](-\frac{\norm{\vxi}}{r})]\nabla\fF[v_{\rho_1}''](\vxi)\nonumber\\
        &-\frac{\Gamma(d/2)}{2\sqrt{2}\pi^{\frac{d+1}{2}}\sigma_b}\nabla\cdot(\Exp_{a,r}[\frac{r^2}{\norm{\vxi}^{d-1}}\fF[g_2](\frac{\norm{\vxi}}{r})\fF[g_5](-\frac{\norm{\vxi}}{r})]\nabla\fF[v_{\rho_1}''](\vxi))
    \end{align*}
    \begin{equation*}
        \fL_{4,\va b}[\fF[v_{\rho_2}]](\vxi)=\frac{\Gamma(d/2)}{2\sqrt{2}\pi^{\frac{d+1}{2}}\sigma_b\norm{\vxi}^{d-1}}
        \Exp_{a,r}[\frac{1}{r}\fF[g_1](\frac{\norm{\vxi}}{r})\fF[g_1](-\frac{\norm{\vxi}}{r})]\fF[v_{\rho_2}](\vxi)
    \end{equation*}
    \begin{equation*}
        \fL_{4,\vw}[\fF[v_{\rho_2}]](\vxi)=\frac{\Gamma(d/2)}{2\sqrt{2}\pi^{\frac{d+1}{2}}\sigma_b}\nabla\cdot(\Exp_{a,r}[\frac{1}{r\norm{\vxi}^{d-1}}\fF[g_2](\frac{\norm{\vxi}}{r})\fF[g_2](-\frac{\norm{\vxi}}{r})]\nabla\fF[v_{\rho_2}](\vxi))\nonumber
    \end{equation*}

    By defining $\frac{\Gamma(d/2)}{2\sqrt{2}\pi^{(d+1)/2}\sigma_b}$ and $h(r^\alpha, g_i, g_j)=\Exp_{a,r}[r^\alpha\fF[g_i](\frac{\norm{\vxi}}{r})]\fF[g_j](-\frac{\norm{\vxi}}{r})]$, finally we have

    \begin{align}
        \fL_{\Delta}[\fF[v_\rho]] &= \fL_{1,\va b}[\fF[v_{\rho_1}'']] + \fL_{1,w}[\fF[v_{\rho_1}'']] + \fL_{2,\va b}[\fF[v_{\rho_2}]] + \fL_{2,w}[\fF[v_{\rho_2}]]\\
        &=\frac{\kappa}{\norm{\vxi}^{d-1}}h(r^3,g_3,g_3)\fF[v''_{\rho_1}](\vxi)+\frac{\kappa}{\norm{\vxi}^{d-1}}h(r,g_4,g_4)\fF[v''_{\rho_1}](\vxi)\nonumber\\
        &+\nabla\cdot(\frac{\kappa}{\norm{\vxi}^{d-1}}h(r^2,g_5,g_4))\fF[v''_{\rho_1}](\vxi)-\frac{\kappa}{\norm{\vxi}^{d-1}}h(r^2,g_4,g_5)\nabla\fF[v''_{\rho_1}](\vxi)\nonumber\\
        &-\nabla\cdot(\frac{\kappa}{\norm{\vxi}^{d-1}}h(r^3,g_5,g_5)\nabla\fF[v''_{\rho_1}](\vxi))\nonumber\\
        &+ \gamma\frac{\kappa}{\norm{\vxi}^{d-1}}h(r,g_3,g_1)\fF[v_{\rho_2}](\vxi)
        -\gamma\frac{\kappa}{\norm{\vxi}^{d-1}}h(1,g_4,g_2)\nabla\fF[v_{\rho_2}](\vxi)\nonumber\\
        &-\gamma\nabla\cdot(\frac{\kappa}{\norm{\vxi}^{d-1}}h(r^2,g_5,g_2)\nabla\fF[v_{\rho_2}](\vxi)),\\
        \fL[\fF[v_\rho]] &= \fL_{3,\va b}[\fF[v_{\rho_1}'']] + \fL_{3,w}[\fF[v_{\rho_1}'']] + \fL_{4,\va b}[\fF[v_{\rho_2}]] + \fL_{4,w}[\fF[v_{\rho_2}]]\\
        &=\frac{\kappa}{\norm{\vxi}^{d-1}}h(r,g_1,g_3)\fF[v''_{\rho_1}](\vxi)
        -\frac{\kappa}{\norm{\vxi}^{d-1}}h(1,g_2,g_4)\nabla\fF[v''_{\rho_1}](\vxi)\nonumber\\
        &-\nabla\cdot(\frac{\kappa}{\norm{\vxi}^{d-1}}h(r^2,g_2,g_5)\nabla\fF[v''_{\rho_1}](\vxi))\nonumber\\
        &+\gamma\frac{\kappa}{\norm{\vxi}^{d-1}}h(\frac{1}{r},g_1,g_1)\fF[v_{\rho_2}](\vxi)
        -\gamma\nabla\cdot(\frac{\kappa}{\norm{\vxi}^{d-1}}h(\frac{1}{r},g_2,g_2)\nabla\fF[v_{\rho_2}](\vxi))
    \end{align}

\end{appendix}

\end{document}